# Interpretability and Accessibility of Machine Learning in selected Food Processing, Agriculture and Health Applications

N Ranasinghe[1], A Ramanan[2], S Fernando[3], PN Hameed[4], D Herath[5], T Malepathirana[1], P Suganthan[6,8], M Niranjan[7] and S Halgamuge[1]

[1] Department of Mechanical Engineering, University of Melbourne, Australia
[2] Department of Computer Science, Faculty of Science, University of Jaffna, Sri Lanka.
[3] Department of Computational Mathematics, Faculty of Information Technology, University of Moratuwa, Sri Lanka.
[4] Department of Computer Science, Faculty of Science, University of Ruhuna, Sri Lanka.
[5] Department of Computer Engineering, Faculty of Engineering, University of Peradeniya, Sri Lanka.
[6] School of Electrical Electronic Engineering, Nanyang Technological University, Singapore.
[7] Department of Electronics and Computer Science, Faculty of Engineering and Physical Sciences, University of Southampton, United Kingdom.
[8] KINDI Center for Computing Research, College of Engineering, Qatar University, Doha, Qatar.

## SUMMARY

Artificial Intelligence (AI) and its data-centric branch of machine learning (ML) have greatly evolved over the last few decades. However, as AI is used increasingly in real world use cases, the importance of the interpretability of and accessibility to AI systems have become major research areas. The lack of interpretability of ML based systems is a major hindrance to widespread adoption of these powerful algorithms. This is due to many reasons including ethical and regulatory concerns, which have resulted in poorer adoption of ML in some areas. The recent past has seen a surge in research on interpretable ML. Generally, designing a ML system requires good domain understanding combined with expert knowledge. New techniques are emerging to improve ML accessibility through automated model design. This paper provides a review of the work done to improve interpretability and accessibility of machine learning in the context of global problems while also being relevant to developing countries. We review work under multiple levels of interpretability including scientific and mathematical interpretation, statistical interpretation and partial semantic interpretation. This review includes applications in three areas, namely food processing, agriculture and health.

# INTRODUCTION

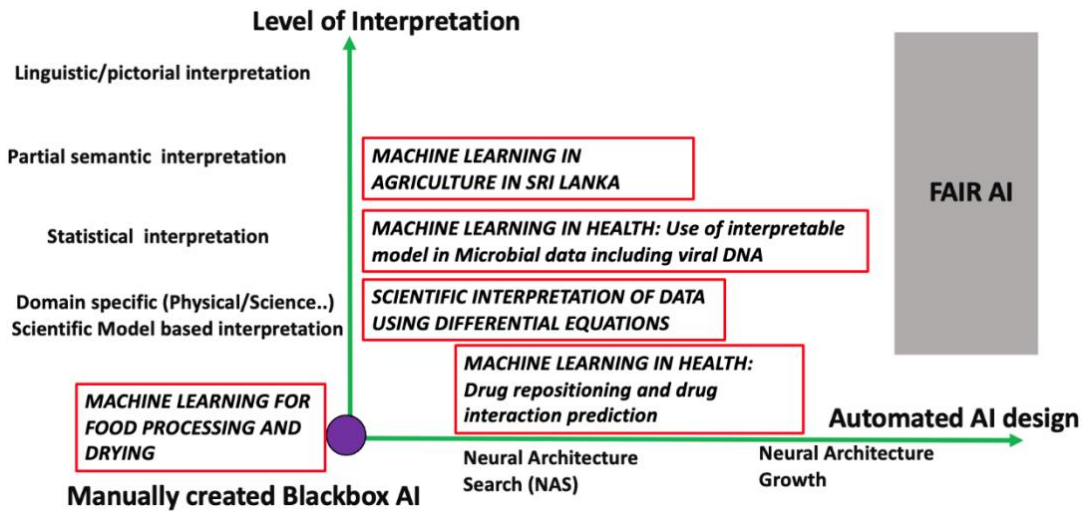

Figure 1 – Contributions of the Paper and Fair, Accessible, Interpretable and Reproducible (FAIR) AI adapted from (Halgamuge S., 2021)

Artificial intelligence (AI) has seen an explosive growth over the last 20 years, largely through recent advances in machine learning (ML) – the data-centric branch of AI. A data-centric AI system consists of an AI model (a structure or architecture) and a method or learning algorithm that enables that model to derive usable information from data. Sometimes the data are exploratory, like the genomic data arriving from different parts of the world about constantly mutating viruses. To discover the presence of new variants or labels, we can feed an AI model with such uninterpreted data, so that researchers will be able to use this AI model to assign labels. Such AI models need unsupervised learning (UL) algorithms to extract information from unlabeled and uninterpreted data. We could also ask those researchers themselves to label data with appropriate variant labels, and feed both labels and genomic data to an AI model that can then use a supervised learning algorithm like deep learning (DL), so that it can serve as a predictor for known variants of the virus. If such an AI model of sufficient strength requires it to be large, deep and complex, we call it a deep neural network (DNN). Shallow neural networks, commonly referred to as Neural Networks (NNs) are data driven mathematical models consisting of about three layers of artificial neurons or nodes (several linear and nonlinear processing elements) which are interconnected through weighted connections.

Popular models of AI, in particular, ML based models have significant deficiencies preventing its broader application. For example, they are mostly uninterpretable (Halgamuge S. , 2021). This paper addresses some of these in the context of several major global problems in the post-pandemic world with relevance to developing countries.

Achieving interpretability in ML will promote broader and effective use of ML, by answering core questions about traceability, accountability, ethical compliance, and inherent biases. Serious concerns remain about the reliability of ML systems in certain contexts. For example, the claim about achieving self-driving status for vehicles has not allayed public concerns. Explainable AI (XAI) methods can help in some applications although may not provide a full explanation about the decision process of the ML model.

There are two main strategies that exist for achieving higher accessibility through automated ML model design. In "growing the ML model from scratch" the ML model starts with a default simple model and grows until its capacity is enough to solve the problem. This approach has been used in both supervised and unsupervised learning-based ML systems. In the "search for the best among select candidates" approach also called neural architecture search (NAS) the best ML model is selected from a list of candidate solutions. The former is applicable to both labelled and unlabeled data, whereas the latter is generally confined to labelled data.

Most ML research on increasing interpretability of ML systems on the Y-axis and accessibility through increased AI model design automation on the X-axis (Figure 1) lies along or close to the axes. Research into interpretable neural networks designed with minimal expert intervention (FAIR AI) will increasingly close a significant knowledge gap, also informed by relevant studies for example, ML with continuous and life-long learning capability (Senanayake, Wang, Naik, & Halgamuge, 2021).

This paper is organised with an example of scientific interpretation capability of ML models using differential equations followed by relevant applications of ML in three areas of importance namely Food Processing, Agriculture and Health: Examples of uninterpretable ML models applied to food drying, partially interpretable Convolutional Neural Networks (CNN) and XAI in plant disease detection including in rice cultivation, safety of taking multiple pharmaceutical drugs and reuse of existing drugs for new diseases using semi-automated unsupervised ML model construction and shedding some light into yet largely unexplored world of microbes including viruses using semi-automated ML model construction.

**SCIENTIFIC INTERPRETATION OF DATA USING DIFFERENTIAL EQUATIONS**

Supervised learning problems generally focus on learning a relationship that maps a given input space to an output space based on input-output pairs. This involves training a model to learn this relationship by looking at a labelled dataset. Although most trained models including neural networks can be represented as a mathematical function, interpreting this function would typically be quite challenging. Only a very few models such as linear regression models will have an interpretable mathematical equation. Convolutional Neural Networks (CNN) are ML models with some (pictorial) interpretability. It can be noted that most equations modelling real-world problems do not have a very large number of terms (Brunton, Proctor, & Kutz, 2016). Furthermore, differential equations (DE) are commonly seen as governing equations of dynamical systems. These DEs are generally derived using first principles. Recently there has been a focus on recovering governing differential equations from observation data of dynamical systems.

(Udrescu & Tegmark, 2020) propose a recursive multi-step method to search through the possible space of equations that fit a given dataset. It uses properties commonly found in real-world physics equations to reduce the search space. These include symmetry, separability, compositionality and more simplifying properties. This paper uses a neural network as a function approximator to discover some of these simplifying properties. The final equation is discovered using a brute force search across the simplified solution space. (Udrescu, et al., 2020) improve upon this approach to discover Pareto-optimal formulae (complexity vs accuracy).

(Brunton, Proctor, & Kutz, 2016) approach the same problem by using sparse regression techniques to circumvent the problem of searching through a large space of possible solutions. Here, the differential equation is assumed to be in the form of,

$$\frac{d}{dt}X(t) = f(X(t))$$

Where, X is the independent variable and f(X(t)) is assumed to be a linear combination of non-linear functions of X. A library $\theta(x)$ of possible terms are constructed using these non-linear functions. The resultant formulation is,

$$\frac{d}{dt}X(t) = \theta(X)\mathrm{E}$$

Where E includes the coefficients corresponding to each library term. The problem is now in the form of a linear regression problem. The paper uses a sparse regression algorithm to solve for the coefficients since real life differential equations typically do not have many terms in the right-hand side. This idea is improved and applied to learn partial differential equations (PDE) in (Rudy, Brunton, Proctor, & Kutz, 2017). (Zhang & Lin, 2018) propose a method to use Bayesian sparse regression to increase the robustness of the learned equation and quantify the uncertainty of the solution.

Neural networks are also used in recent work to recover governing differential equations from observed data. (Martius & Lampert, 2016) use neural network algebra to learn equations through backpropagation. A shallow neural network is used with custom activation functions which include multiplications, trigonometric functions and identity functions. Sparsity promoting L1 regularization is used to promote learning a simple equation with a low number of terms. (Sahoo, Lampert, & Martius, 2018) proposes a method to extend the class of the learnable equations using an equation learning network to include divisions. (Long, Lu, & Dong, 2019) propose a deep neural network architecture to discover time dependent PDEs from observed data. This architecture proposes the use of a $\delta$-block, which uses convolutions to approximate differential operators and a symbolic neural network to approximate the non-linear response function. The approximation framework is as follows:

$$\widetilde{U}(t + \delta t) \approx \widetilde{U}(t) + \delta t * F$$

where, $\widetilde{U}(t)$ is the predicted value at time t and F is the PDE-NET approximation. The architecture of a $\delta$-block is shown in Figure *2*. A single $\delta$-block can only approximate one-step dynamics, meaning that it is prone to error accumulation over time. Multiple $\delta$-blocks with shared parameters are stacked to approximate multi-step dynamics and reduce error accumulation.

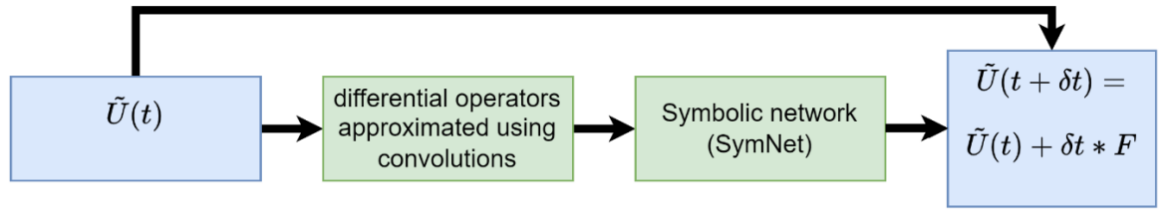

Figure 2- PDE-NET 2.0 *(Long, Lu, & Dong, 2019)*

Most works in this area focus on learning a certain type of ordinary differential equation (ODE) or PDE. However, when applying these methods to real scenarios, we may not know the type of equation in advance. Furthermore, most of the approaches in the literature can only learn $1^{st}$ order differential equations. Therefore, future work could focus on learning higher order DEs. Scalability is also a challenge in current methods and is especially evident in NN based equation learning algorithms. These are quite resource intensive and in certain cases require GPUs to run. Moreover, only a very few works explore methods of uncertainty quantification of the solution. This is also a potential research direction.

**MACHINE LEARNING FOR FOOD PROCESSING AND DRYING**

Food security is a major concern around the world as foods have high nutritional value and are an essential part of everyday life. However, most plant-based foods are highly perishable and about one-third of global food production is lost annually due to inadequate processing (Vilariño et al., 2017). This issue is amplified in developing countries where about 40% of seasonal fruits and vegetables are wasted (Karim & Hawlader, 2005). Drying is a major food preservation and processing technique which aims to remove moisture, preventing microbial spoilage (Kumar et al., 2015), thus adding value to a product, permitting early harvest, reducing shipping weights and costs, minimizing packaging requirements and increasing shelf-life (Zielinska et al., 2013). However, drying is highly complex involving interconnected simultaneous momentum, heat and mass transfer with time-varying physiochemical and anisotropic structural changes which depend on the dynamic product-drying environment interactions occurring (Khan et al., 2020; Welsh et al., 2018). Additionally, the underlying mechanisms, which originate within the materials cells, are still not well understood. Due to these complexities, optimising the design of dryers and process conditions is challenging and expensive. AI, for example uninterpretable ML, has demonstrated great potential to facilitate innovation and optimisation for the processing of foods, providing a low-cost alternative to the current design modelling approaches and in-line process controllers. Converting these ML models to interpretable ML models will have significant advantages.

**Uninterpretable Machine Learning**

ML (e.g. NN) is able to investigate, model and predict the nonlinear time-varying behaviour of food material during drying (Angermueller et al., 2016). The three main parts of NN represent the input layer (material definition and/or process conditions), the hidden layer (consisting of nodes and weighted connections) and the output (drying rate, moisture content evolution and/or quality). The approach utilises experimental examples of the system which is being modelled (training data) to optimize the parameters including weights of the connections between neurons to predict the complex outputs. Over the years various NN models have been applied in food drying for predicting complex outputs for various materials (Sun et al., 2019) and various drying technologies (Sarkar et al., 2020). Generally, ML is applied to predict four types of outputs for food processing, the evolution of moisture and temperature, morphological/structural changes, transfer coefficients and quality changes.

Common complex outputs predicted through ML models include the moisture and temperature evolution of the material and the drying rate/transfer coefficients of a dryer. Predicting such complex outputs can provide insight into the drying time and the energy consumption of a dryer. Chasiotis et al. (2020) utilised a NN to predict the moisture content evolution of convective drying cylindrical quince slices. This work utilised 1372 experimental samples split between training data and cross validation data to consider three input variables (temperature, flow velocity and time) to predict the samples' moisture content. The results showed good agreement between the predicted and experimental values. Çerçi and Daş (2019) applied a NN and decision tree to predict the heat transfer coefficient for natural and force convection, concluding NN was more successful in estimating the heat transfer coefficient. Saraceno et al. (2012) investigated and compared three different modelling approaches, a thin-layer model, a NN and a hybrid neural model for two different vegetables with different characteristic dimensions for a wide range of process conditions. Their work demonstrated pure neural models gave very accurate predictions when modelling/reproducing known data/process conditions. However, the accuracy of the NN model significantly decreased when attempting to extrapolate/apply the model to unknown scenarios/process conditions.

The morphological and quality changes food material experiences during drying have also been predicted through ML techniques. Scala et al. (2013) constructed a NN model for predicting the quality characteristics of Granny Smiths apples during convective dehydration. The work investigated experimental data from three different drying temperatures (40, 60 and 80 °C) drying at three air flowrates (0.5, 1.0 and 1.5 m/s) and effectively predicted the colour, water capacity and total phenolic content of the samples. Additionally, the work identified the optimal drying conditions within the experimental process conditions. Chen and Martynenko (2013) utilised computer vision to evaluate the drying rate, shrinkage and colour changes of two varieties of blueberry, though the accuracy of the shrinkage measurements was limited by the pixel resolution of the camera and the accuracy of the colour measurement was limited by the quality of illumination and colour reproduction. Recently, Sinha and Bhargav (2022) develop an ANN model to predict key properties related to shrinkage, specifically solid density, initial porosity and initial water saturation of a given food material, using temperature and moisture data from a set of simple experiments. The work demonstrates how a NN model can serve as an efficient indirect method of property estimation in food material.

ML techniques have also been applied to calculate key transport properties. Mariani et al. (2008) developed an NN-based inverse method to estimate the apparent diffusivity of bananas at different drying temperatures. The work found a small change in drying temperature and moisture content caused a significant change in bananas' apparent diffusivity. Sablani and Rahman (2003) developed an NN model for predicting the thermal conductivity of various foods in terms of moisture content, temperature and apparent porosity. NN models have demonstrated great predictive capabilities in comparison to other statistical approaches.

Although AI and ML techniques provide a low-cost alternative to facilitate innovation and optimisation for the processing of foods, applying ML to food processing does have some challenges. ANN is considered to be a "black box" approach (i.e. uninterpretable) where the user cannot see what is happening during the simulation which is not ideal for understanding what is occurring during drying. Additionally, ML is a data-driven approach where the accuracy of the model heavily depends on acquiring a large comprehensive training dataset. However, food processing data is often scarce and obtaining large datasets is expensive due to the resources and costs involved in measuring this data experimentally. As a result, most NN models in literature have been constructed using small datasets and therefore their accuracy significantly decreases when applying the model to data outside the training observatory data (Saraceno et al., 2012). Recently, a new class of NN or deep learning, physics-informed neural networks (PINN), has emerged that can seamlessly integrate training data and complex mathematics to optimise a loss function. By incorporating prior knowledge, it minimises the need to have a large observatory training dataset for maintaining the high accuracy of the predictive model. Hence, overcoming the main challenges for AI in food processing. A PINN model can be trained with additional information obtained by enforcing physical laws with mathematical governing equations (Karniadakis et al., 2021). PINN models have been constructed for the deformation of elastic plates (Li et al., 2021), engineering heat transfer applications (Zobeiry & Humfeld, 2021), modelling fluid mechanics (Raissi et al., 2020), understanding permeability and viscoelastic modulus properties (Yin et al., 2021) and to solve forward and inverse problems (Zhang et al., 2019). Though demonstrating great potential, the PINN modelling approach has not been applied to food drying yet. For additional information on AI for the application of food processing, readers are directed to the comprehensive reviews of Sun et al. (2019), Khan et al. (2020) and Nayak et al. (2020). More insights into PINN modelling approaches can be found in Karniadakis et al. (2021).

## MACHINE LEARNING IN AGRICULTURE IN SRI LANKA

Sri Lanka is a tropical country with a high potential for cultivating and processing a variety of crops. Although the open-field agricultural system is the most prominent in Sri Lanka, the sector is increasingly enabled by the recent growth of IoT applications and AI systems in the field. Greenhouse agriculture is an emerging subsector today that attempts to maximize the harvests under a protected cultivation environment within the limited space. The recent government restrictions on chemical fertilizers have regulated the local agriculture industry to look for optimal fertilizer combinations that maximize yield in open-field systems and green-house environments. As the research of different permutated fertilizer combinations increases, numerous plant diseases and nutrient deficiencies receive increased attention from ML researchers.

Rice is one of the most popular food crops in the World and one of the main grains used in Asia including in Sri Lanka. Jaffna is located in the Northern tip of Sri Lanka at a longitude of 79° 45' - 80° 20' and east latitude of 9° 30' - 9° 50' with the population of around 700,000. The agriculture and fishery sectors play a crucial role in the gross production of Jaffna. Since the soil and climatic conditions are favourable to cultivate a wide range of crops including paddy, agriculture plays a significant role in the lives of people in Nothern Sri Lanka. Various diseases in paddy agriculture have been seriously affecting rice production and constitute a big challenge for the agricultural community to ensure food security. These diseases include rice blast, bacterial leaf blight, bacterial leaf streak, sheath blight, seedling blight, false smut, rice hispa, sheath rot, root knot, leaf streak, yellow stem borer, brown spot, and brown planthopper (see Figure *3*) mainly caused by viruses, bacteria, fungi etc. Diseases of rice plants could be affected by different factors, such as fertilizers, nutrients, water management, climatic conditions, lighting conditions, humidity, and farming conditions. The detection of such rice plant diseases is normally performed on the visual assessment of the symptoms which is subjective, time-consuming even for well-experienced experts and is prone to error. Automating such visual assessment will provide information for the prevention and control of rice disease through which the quality and the quantity of rice production can be increased by reducing the operations costs. Thus, the automation will contribute significantly to the economic growth of Sri Lanka.

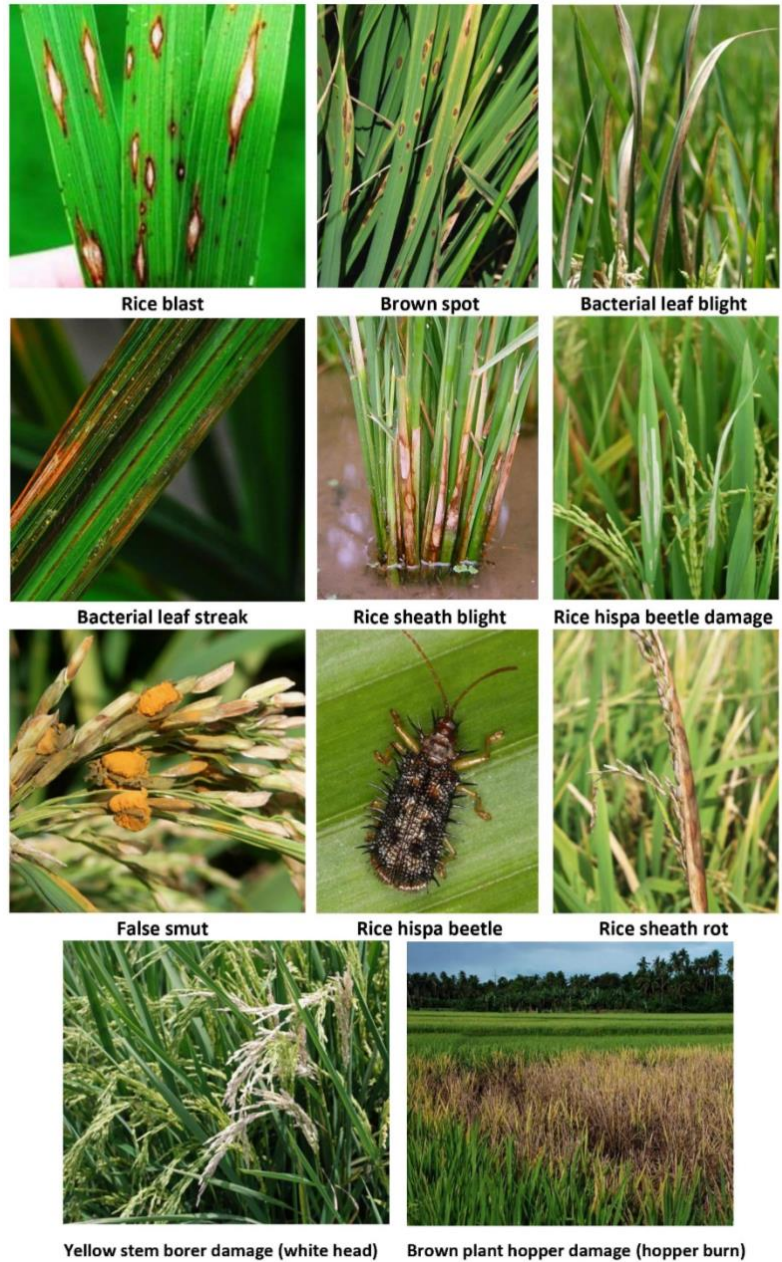

Figure 3- Examples of common rice diseases (courtesy of http://www.knowledgebank.irri.org/ )

**Explainable AI for plant disease classification**

The diagnosis of plant nutrient deficiencies at early stages is essential as it indicates the impact of the applied fertilizer combination on plant growth. This deficiency could be a reason for overdose or lack of a particular nutrient or a combination. These deficiencies could be mainly seen in the plant's stems, leaves, or terminal buds (see Figure *3*). The association of these symptoms to different nutrient deficiencies requires expert knowledge. A data-centric approach is to use a classification model trained on labeled data. Image processing applications in the field of agriculture help to identify and diagnose the crop diseases, pests and weeds, and the classification of agricultural products. Convolutional neural network (CNN) is one of the most effective methods of image processing which extracts image features, achieves high-level fusion of semantics and deep extraction of features through multilayer networks. Due to inherent learning capability, CNN is the prominently used machine learning model for plant disease classification. The convolution operation identifies the local features and through the processing in the subsequent layers of CNN, it hierarchically develops motifs and parts that could represent the symptoms of a particular disease. Despite the high classification accuracy, the CNN architecture is further explored to shed some light into the explanation or interpretation of its decision making. While techniques such as temperature scaling (Guo, Pleiss, Sun, & Weinberger, 2017) have been used to calibrate the CNN confidence scores, the explainable artificial intelligence models (XAI) have been hybridised with the CNN architectures (XAI-CNN) to provide explanations for such classifications. In general, these XAI models relate to or explain the class labeling by identifying the mostly contributed regions of the input image. The more commonly used XAI techniques are reviewed next emphasizing the main principle behind them.

Generally, XAI techniques that explain single disease predictions are of great interest in plant disease classification. CAM (Zhou, Khosla, Lapedriza, Oliva, & Torralba, 2016), LIME (Ribeiro, Singh, & Guestrin, 2016), and SHAP (Lundberg & Lee, 2017) are commonly used XAI models which provide justifications by generating local attention maps or by modeling feature importance. CAM (Class Activation Map) and its derivatives such as GRAD-CAM (Selvaraju, et al., 2017) and GRAD-CAM++ (Chattopadhay, Sarkar, Howlader, & Balasubramanian, 2018) are visual explanation techniques that can identify discriminative image areas by locating the pixel groups responsible for influencing the association between a given input image and a particular class label. These pixels are determined by projecting classification layer weights or classification layer "gradients of activation" to the convolution feature maps. In contrast, LIME attempts to develop linear (interpretable) models over the complex perceptron model by identifying significant features of a data sample. It modifies a single data sample by tweaking the feature values (or identifying super-pixels) to understand the relationship between input and the predicted output. By doing so, LIME provides a list of explanations reflecting the contribution of each feature in the prediction of the disease class of a given data sample. In SHAP (Shapely Additive Explanations), the feature values of a data sample act as players in a coalition. Shapley value is the average marginal contribution of a feature value. These values calculate the feature importance and explain the association of the data sample to a particular class label. When applying these XAI models, several challenges are recognized. In LIME, it is hard to model non-linear relationships with linear models even by tweaking features or identifying the super-pixels. The perturbations of a single data point with random noises would not cover the variations in the dataset. In SHAP, the Shapley values provide additive contributions of explanatory features, which might be misleading if the model is not additive. Further, it is not mathematically feasible to calculate the Shapley values for all the classification problems. Therefore, mainly CAM-based XAI techniques have been applied in plant diseases classification (Liu, et al., 2020) (Ennadifi, Laraba, Vincke, Mercatoris, & Gosselin, 2020) demonstrating significant improvements in the explainability of the class-label predictions.

When employing CNN together with XAI models, few challenges are recognised. First, publicly available datasets are small and often imbalanced leading to models getting overfitted even if the data are augmented with various augmentation techniques. Therefore, when applying XAI techniques on small and imbalanced datasets, CAM and GRAD-CAM approaches do not perform well because it becomes difficult to locate the correct region of the convolution feature maps just by activating the classification layer of these models. This problem gets severe as the number of classes increases. Therefore, developing large balanced datasets is vital.

Second, a deficiency could be associated with multiple stages: early-stage, mild-stage, and severe stage and symptoms of several deficiencies could be overlapped and making the classification challenging. For example, there could be both Ca+ and B deficiency; their symptoms could not be separable at their early stages but separable when they reach the mild stage. Therefore, applying CNN models at different stages of the deficiency could result in different predictions with a low confidence score. Using XAI models on poorly calibrated models further decays the explanations' accuracy. Therefore, reasons based on feature activations or feature importance are insufficient for plant disease classification. It requires a much richer analysis of the features to map their activation to more probable two or three classes (including the predicted) based on the stages of the deficiency and growth stage of the plant.

Although existing XAI techniques give explanations in terms of feature activation and importance that can be interpreted by machine learning researchers they cannot fully explain the decision-making process in terms of agricultural technology and therefore of less use in real applications in disease recognition in agriculture. Therefore, these illustrations must map into human-understandable explanations for example by integrating the concepts of expert systems on the activated convolution features of XAI.

**Rice disease detection using Convolutional Neural networks**

Due to the inherent ability of the CNN model to show layer by layer pictorial explanation of how an input image is processed to reach the classification outcome, it is a good candidate to use in detecting rice diseases even without hybridizing with XAI.

(Jiang, Lu, Chen, Cai, & Li, 2020) proposed a rice disease recognition model that combines CNN and support vector machine (SVM) to identify four types of rice diseases. Images are pre-processed by applying mean-shift algorithm to segment lesions from rice leaf disease. From the images of leaf lesions shape features such as area, roundness and shape complexity were extracted, and the colour features are extracted using CNNs by converting RGB colour space into HSI and YCbCr colour space. The extracted features are then classified by an SVM with RBF kernel due to the small number of sample images. They collected a total of 8911 images from rural farmland and from rice leaf disease atlas in which 6637 images of disease rice were manually cropped. A classification rate of 96.8% is reported.

(Anandhan & Singh, 2021) proposed a rice disease recognition model by comparing Mask R-CNN and Faster R-CNN algorithms to identify five types of rice diseases. Images are pre-processed using a multistage median filter to reduce noise. They implemented the model using ResNet-50 (Residual Neural Network) as the backbone and optimised the model using a stochastic gradient descent optimiser. In each iteration, a mini-batch of size 60 was used. The learning rate is set to 0.01 with weight decay fixed to 0.0001. They collected about 1500 images from rice plants in the region under different weather conditions and images were manually annotated for diseases. A classification rate of 87.5% is reported for Mask R-CNN which outperforms Faster R-CNN.

(Liang, Zhang, Zhang, & Cao, 2019) proposed a rice blast recognition method using CNN. In their work, high-level features extracted by CNN were compared with handcrafted features: Harr-wavelet Transform and local binary patterns histograms (LBPH) in rice blast recognition. In addition to using CNN in the recognition task, the extracted features were also classified by SVM using an RBF kernel. Moreover, they created a rice blast disease image set that is made publicly available†. Images encompassing 2902 negative and 2906 positive samples were considered for testing and training a CNN. Quantitative analysis results show that CNN with Softmax and CNN with SVM have almost the same performance. A classification rate of 95.83% is reported for CNN showing better performance for the said recognition task.

†http://www.51agritech.com/zdataset.data.zip

(Zhou, Zhang, Chen, He, & Ma, 2019) proposed an approach to identify three types of rice diseases based on FCM-KM and Faster R-CNN fusion techniques. Images are pre-processed using a weighted multilevel median filter for noise removal. Thereafter, Otsu threshold segmentation algorithm (Otsu, 1979) is applied to segment lesions from leaf disease images. In order to extract rice disease characteristics, the FCM-KM is used to reset the bounding box size in Faster R-CNN for the convergence to be accelerated. The optimal value of K in K-means is selected with the maximum and minimum distance algorithm and where those initial cluster centres should be positioned. In addition, the dynamic population firefly algorithm based on the chaos theory is applied to the clustering process to jump out of the local optimum and obtain faster convergence. The FCM-KM and Faster R-CNN was applied to extract the disease characteristics and classify the images for pests and diseases. An image set consisting of 3010 images were tested and a classification rate of 97.2% is reported for the proposed method.] is applied to segment lesions from leaf disease images. In order to extract rice disease characteristics, the FCM-KM is used to reset the bounding box size in Faster R-CNN for the convergence to be accelerated. The optimal value of K in K-means is selected with the maximum and minimum distance algorithm and where those initial cluster centres should be positioned. In addition, the dynamic population firefly algorithm based on the chaos theory is applied to the clustering process to jump out of local optimum and obtain faster convergence. The FCM-KM and Faster R-CNN were applied to extract the disease characteristics and classify the images for pests and diseases. An image set consisting of 3010 images was tested and a classification rate of 97.2% is reported for the proposed method.

(Sreevallabhadev, 2020) proposed a method to identify rice blast disease by using CNN for feature extraction and SVM for classification. Experiments were carried out with 60 experimental configurations that vary in the choice of deep learning architecture, training mechanism, and dataset type. They used approximately 60,000 images of plant leaves in their original dataset‡ and created three different versions of the dataset by considering colour, grayscale and leaf segmented images. An AlexNet based in CNN is used in extracting the features that are then fed to an SVM classifier. A classification rate of 96.8% is reported.

‡https://www.kaggle.com/abdallahalidev/plantvillage-dataset

It is worth noting that, a majority of the work reported in the literature uses the following experimental setups and points out that there is a huge room for carrying out research in this domain for the betterment of the economy using deep learning approaches.

- Evaluate on small scale datasets consisting of 500 to 5000 images except the work in (Sreevallabhadev, 2020). The small sized datasets directly impact the quality of the mapping function approximated by neural networks

- Develop models to recognise a few rice diseases though there are several known diseases that greatly affect the quality and quantity of rice production

- Propose methods mainly using features such as shape, texture and colour by the use of histogram-oriented gradient (HOG), scale-invariant feature transform (SIFT), wavelet, local binary pattern (LBP), HSI and YCbCr, CNN, etc.

**MACHINE LEARNING IN HEALTH**

**Drug repositioning and drug interaction prediction**

Drug repositioning and drug interaction prediction are two fundamental applications of drug development and clinical care that have significant benefits in pharmacology. Repositioning of existing drugs can be classified as single drug-based repositioning and drug combination-based repositioning. Drug combinational treatments are identified to be much effective for treating many diseases. Moreover, drug-drug interactions (DDIs) are likely to occur when a pair of drugs or a combination of drugs are co-administered. Investigating harmful DDIs is essential to enhance the effects of clinical care.

In these contexts, the features of existing drugs, such as chemical structures, gene expressions, target proteins, side effects, indications, etc., are considered to compare the pairwise drug similarity (Sun, Hameed, Verspoor, & Halgamuge, 2016) (Hameed P. , Verspoor, Kusljic, & Halgamuge, 2018) (Hameed P. , Verspoor, Kusljic, & Halgamuge, 2017) Further, chemical structural data, gene expression data, side effect related data and the transcriptional responses have been used for predicting drug mechanism of action (MoA), i.e., the prediction of molecular targets for a particular drug. "Connectivity Map" (Lamb, et al., 2006) resource, which is frequently used in related studies, can be used to find connections among small molecules sharing an MoA, chemicals and physiological processes, and diseases and drugs.

Over the past two decades, machine learning-based and network analysis-based drug repositioning have gained popularity. Machine learning approaches incorporate clustering, classification and deep learning techniques with statistical concepts, whereas network analysis approaches represent pharmacological knowledge as networks with nodes and edges. In network-based analysis, features such as genes, proteins, molecules and phenotypes can be used as the nodes. Their functional similarities, mode of actions and relationships can be represented by the edges which interconnect the nodes.

In network-based analysis, techniques such as Random walk (Zhang, et al., 2017), Node embedding (Su, et al., 2021), Matrix Perturbation (Zhang, et al., 2017), and Steiner-tree based algorithms (Sun, Hameed, Verspoor, & Halgamuge, 2016) were used. They focus on building a network using drug features, actions, and characteristics to predict the novel drug-repositioning candidates via drug-drug, disease-disease, and drug-disease relationships. (Zhu, et al., 2020), proposed a drug-centric graph model, extracted and integrated six drug knowledge bases and constructed the drug knowledge graph. They have used a path-based data representation method and embedding-based data representation for comprehensive analysis for drug repositioning. Both methods applied to the drug knowledge graph evidenced better predictive performance on diabetes mellitus treatments. In Matrix Perturbation, novel drug-disease prediction can be transformed into a missing link prediction problem. In (Sun, Hameed, Verspoor, & Halgamuge, 2016), the Prize-Collecting Steiner Tree approach has proven to be a promising subnetwork identification method for drug repositioning. Their Physarum-inspired subnetwork identification algorithm employed on drug similarity networks has inferred useful repositioning candidates for cardiovascular diseases.

Positive Unlabeled Learning (PUL) is an emerging topic in the field of computational drug repositioning and drug interaction prediction as they involve positive and unlabeled. The application of PUL for single-drug repositioning (Mordelet & Vert, 2011) (Yang, Li, Mei, Kwoh, & Ng, 2012) and DDI prediction (Hameed P. , Verspoor, Kusljic, & Halgamuge, 2017) (Zheng, et al., 2019) has shown improved predictive performance. PUL enables prioritizing plausible negatives from the unlabeled data and improves performance compared to randomly selecting negatives from the unlabeled data. (Mordelet & Vert, 2011) introduced a scoring function and assigned a specific score to each data pair through which the data are sorted in descending order to distinguish positives and negatives. Similarly, (Yang, Li, Mei, Kwoh, & Ng, 2012) have classified unlabeled data as reliable negatives, likely positives, likely negatives and weekly negatives. (Hameed P. , Verspoor, Kusljic, & Halgamuge, 2017) also identified likely negatives from the unlabeled DDI pairs to treat as negatives when developing the binary classification models. These PUL approaches have shown significant improvements in final predictions.

Drug combination-based repositioning is emerging research in computational drug repositioning. Recent studies focused on computational drug combination-based repositioning, exploring both therapeutic uses and adverse effects of drug combinations. Since there exist approximately 16,000 approved drugs on the market (Wishart, et al., 2018), millions of drug combinations can be formed. However, only a small number of drug combinations are confirmed with experimental research. Therefore, there is a need for accurate and reliable approaches to infer repositioning candidates from those millions of unlabeled drug combinations. PUL can produce significant improvements in drug combination-based repositioning as well. Further, awareness of harmful DDIs is beneficial to extract the most suitable drug combinations for drug repositioning.

Imbalanced data is another challenge arising in the drug repositioning and drug interaction prediction domains. Since the positive: unlabeled ratio is significantly high, balanced training sets are frequently used in machine learning approaches in these contexts. The use of balanced datasets in model training can improve the generalizability of the trained model by preventing the model from being biased towards a particular class. (Wei, Dunbrack Jr, & Roland, 2013) discovered that regardless of the rates of positives and negatives in human genome data, support vector machines trained on balanced data sets have performed well in binary classification. (Leevy, Khoshgoftaar, Bauder, & Seliya, 2018) have also emphasized the issues that may arise with high-class imbalanced datasets and have presented the possible solutions that can be applied to avoid such potential issues. Further, (Hameed P. , Verspoor, Kusljic, & Halgamuge, 2017) emphasized using balanced datasets and employed multiple balanced training sets to strengthen the final prediction using ensemble learning which can reduce the variance of the final outputs.

**Use of interpretable models in Microbial data including viral DNA**

Microbiology is the study of the structure and functions of microorganisms and their interactions with other microorganisms, species, or environments (DiMucci, Kon, & Segrè, 2018) (Xie, et al., 2019) (Moitinho-Silva, et al., 2017). Due to the advancements in microbial sequencing technologies, for instance, 16S rRNA sequencing, microbial studies generate a massive amount of data with a large number of samples and variables. Since most microorganisms are not present in most samples, the microbial data are also known to be highly sparse (Martino, et al., 2019), adding to its challenges. Therefore, ML has been employed as a powerful tool that can analyze and identify significant patterns in microbial communities (Ghannam & Techtmann, 2021).

ML algorithms can appear in many forms in microbial data analysis. Dimensionality reduction and visualization leading to better interpretation of data is one such significant manifestation. To better comprehend the underlying patterns in the high dimensional microbial data, it is necessary to reduce the number of dimensions either to visually interpretable two or three dimensions or to a reasonable number of dimensions that can be used as the input to another ML model. Principal Component Analysis (PCA) (Wold, Esbensen, & Geladi, 1987) and Principal coordinate analysis (PCoA) (Kruskal & Joseph, 1978) are two techniques commonly used for the task. However, they fail to capture the highly non-linear relationships in some microbiome data (Xu, Schultz, & Xie, 2016). In contrast, manifold learning techniques such as Isomap (Tenenbaum, Silva, & Langford, 2000), Locally Linear Embedding (LLE) (Roweis & Saul, 2000), and t-Distributed Stochastic Non-linear Embedding (t-SNE) (Van der Maaten & Hinton, 2008) can overcome this limitation and therefore are commonly used in exploratory data visualization. t-SNE was also considered regularly for data visualization in the recent past on account of its capability in revealing the local structure in high-dimensional data, also bringing its attention to microbiome data (Kostic, et al., 2015). Moving forward Xueli et al. (Xu, Xie, Yang, Li, & Xu, 2020) proposed a t-SNE based classification method for compositional microbiome data using Aitchison distances as the conditional probabilities. By this method, authors were able to achieve better classification performance compared with the classifiers built in the original high-dimensional space. However, t-SNE's ability in preserving the global structure of data has been questioned. Therefore, a recently proposed technique, Uniform Manifold Approximation and Projection (UMAP) (McInnes, Healy, & Melville, 2018) has gained popularity due to its capability in preserving the local structure while being superior in capturing the global structure compared to t-SNE. UMAP's usefulness in revealing composite patterns in microbiome data is demonstrated by Armstrong et al. (Armstrong, et al., 2021) with their application of the algorithm on three different microbial datasets. Autoencoders are artificial neural network-based algorithms that learn a compressed representation of the high dimensional input by minimizing its reconstruction error. Recently, autoencoders and their variations (e.g. variational autoencoders) have demonstrated their usefulness in learning meaningful latent features from data (Wang, Huang, Wang, & Wang, 2014), showing potential to be used with microbial data. Therefore, the research in this domain would benefit from the concept of interpretable neural networks when optimally identifying a suited architecture for a particular problem and mapping the process of identifying latent patterns to human-understandable explanations.

Metagenomics which is the analysis of DNA sequences of multiple species is effectively being used for studying microbes, especially viruses (Herath, et al., 2017). A metagenomic sample would consist of a large number of DNA sequence reads of multiple species making it inherently complex to study. Sample preparation including the DNA sequencing, reads assembly, annotation, and analysis of the data can be identified as the key steps in a metagenomic experiment and Machine Learning (ML) models are being used in all the mentioned steps (Krause, et al., 2020). Metagenomic data can be used to infer the microbes' interactions with their host environments, and thereby are considered in microbiome biomarker discovery for disease diagnosis and monitoring. However, due to their complex nature, using metagenomic data in precision medicine as a decision support system demands interpretable models with conciseness and readability by non-experts (Prifti, et al., 2020). Predomics, an ML approach inspired by microbial ecosystem interactions attempts to develop an interpretable and accurate model that can be used to analyze any type of data; especially microbiome data. Its effectiveness has been demonstrated on liver cirrhosis data (Prifti, et al., 2020). Furthermore, the effectiveness of an interpretable machine learning approach has been demonstrated by a meta-analysis of 1042 fecal metagenomic samples. It suggests the use of interpretable models to obtain non-intrusive predictive disease biomarkers for colorectal cancer using metagenomics data of the gut microbiome (Casimiro-Soriguer, Loucera, Peña-Chilet, & Dopazo, 2022). Furthermore, another recent study suggests that easily accessible microbiome samples and their metagenomic analysis using interpretable models have the potential for the non-invasive diagnosis of diseases (Carrieri, et al., 2021). The mentioned study demonstrates it by generating explanations for the predictions of skin hydration, age, menopausal status, and smoking habits based on leg skin microbiome (Carrieri, et al., 2021). As such, as the sequencing cost is becoming low, low and middle-income countries can benefit from interpretable models for metagenomics data analysis that can be used for disease surveillance and anti-viral therapy.

## DISCUSSION

Achieving interpretability in AI will promote broader and effective use of AI, by answering core questions about traceability, accountability, ethical compliance, and inherent biases. Several applications of AI reviewed in the paper provide different levels of interpretability from non to explainable AI. Interpretability can be further expanded to achieve linguistic interpretation, e.g.

(Cao, et al., 2020) (Halgamuge S. , 1997).

AI models have narrow accessibility as they are generally manually designed by experienced AI experts, which limits access for fields in which AI experts have no knowledge or interest. Some automation in the design of AI is used in the drug repositioning and drug interaction prediction application described. Other AI methods, for example, evolutionary algorithms can be used to design AI systems automatically (Cao, et al., 2020). There is vast untapped potential: for medical breakthroughs, distribution and use of scarce resources, forecasting with unheard-of accuracy in natural and economic domains, etc when both interpretability and accessibility of AI systems can be combined as shown for small scale neural networks in (Halgamuge S. , 1997).

## ACKNOWLEDGMENT

Authors acknowledge the contribution of A/Prof Azharul Karim and Dr Zachary Welsh on Food Processing section and various grants including ARC DP220101035 and ARC DP210101135, University of Melbourne Scholarships of PhD students Nisal Ranasinghe and Tamasha Malepathirana.